\documentclass{article}
\usepackage{subcaption}

\usepackage{amsmath,amssymb,amsfonts}
\usepackage{algorithmic}
\usepackage{graphicx}
\usepackage{textcomp}
\usepackage{xcolor}
\usepackage{fvextra}
\usepackage{listings}
\usepackage[T1]{fontenc}
\usepackage{lmodern} 
\usepackage{booktabs}
\usepackage[preprint]{corl_2026} 

\hypersetup{hidelinks}

\newcommand{\taskthumb}[1]{%
  \includegraphics[
    width=0.322\linewidth,
    height=0.68in,
    keepaspectratio
  ]{#1}%
}

\newcommand{\taskblock}[4]{%
  \begin{minipage}[t]{0.49\linewidth}
    \raggedright
    {\scriptsize\bfseries #1\par}
    \vspace{1pt}
    \makebox[\linewidth][l]{%
      \taskthumb{#2}\hfill
      \taskthumb{#3}\hfill
      \taskthumb{#4}%
    }
  \end{minipage}%
}

\title{Closing the Loop in Humanoid VLA: Persistent 3D Object Tokens for Verifiable Loco-Manipulation}

\author{
  {\bfseries
  Peng Ren\textsuperscript{1,2,4}\thanks{\normalfont Equal contribution.}\thanks{\normalfont This work was conducted during an internship at DeepCybo.}
  \quad
  Haoyang Ge\textsuperscript{3,2}\footnotemark[1]
  \quad
  Jiang Zhao\textsuperscript{1}
  \quad
  Cong Huang\textsuperscript{2,5}
  \quad
  Yukun Shi\textsuperscript{2,5}
  }
  \\
  {\bfseries
  Pei Chi\textsuperscript{1}\thanks{
    \normalfont
    Corresponding authors:
    Pei Chi (\texttt{peichi@buaa.edu.cn}) and
    Kai Chen (\texttt{kaichen@zgci.ac.cn}).
  }
  \quad
  Kai Chen\textsuperscript{2,4,5}\footnotemark[2]
  }
  \\[0.7em]
  {\bfseries\small
  \textsuperscript{1}BUAA
  \quad
  \textsuperscript{2}BZA
  \quad
  \textsuperscript{3}TJU
  \quad
  \textsuperscript{4}DeepCybo
  \quad
  \textsuperscript{5}ZGCI
  }
}

\begin{document}
\maketitle


\begin{abstract}
Vision-language-action policies are a promising foundation for general robot control, but long-horizon humanoid loco-manipulation requires the robot to treat task objects as persistent physical entities across movement, contact, occlusion, and recovery. 
We study this problem as object-state divergence: the object state used to condition a whole-body action can differ from the state used to decide whether the action achieved the intended physical relation. 
We propose \emph{Persistent Object Tokenization} (POT), which maintains role-indexed 3D object records from RGB-D observations and converts them into object tokens for a whole-body action expert. 
Instantiated as \emph{POT-VLA}, the same object records condition action generation and support geometric predicate checks, yielding a closed-loop execution system in which object state is both actionable and verifiable. 
On a Unitree G1, POT-VLA improves a matched direct GR00T-N1.7 baseline from 39/80 to 71/80 successes over eight real-world task families. 
In an external Being-0-aligned reference, POT-VLA achieves 44/50 successes on aligned service tasks, compared with the 37/50 success reported by the Being-0 paper. 
The largest gains occur on tasks requiring maintained 3D relations, suggesting that persistent object-centered state is a useful abstraction for verifiable humanoid VLA execution.
\end{abstract}



\section{Introduction}
Natural-language commands such as ``tidy up the desk'' and ``bring me a drink'' require a humanoid robot to ground task objects, move through the scene, manipulate them, and verify progress over repeated short-horizon action chunks~\cite{pi_0,pi05,rt1,rt2,zhao2023chat}. 
The central difficulty is not perception or control alone: the same cup, basket, support surface, or handover partner must remain addressable as the robot walks, reaches, grasps, places, observes failures, and recovers. 
This creates a closed-loop state-representation problem for humanoid VLA: object state must be represented in a form that can guide action generation and later verify the physical consequences of those actions.

We identify \emph{object-state divergence} as a practical failure mode in long-horizon humanoid VLA. 
Modern VLA policies typically condition actions on visual-language features from the current observation, while task progress may be checked by a separate monitor, predicate layer, or language-level state. 
When the object state used to act differs from the object state used to verify, small grounding, grasping, or placement errors can lead to premature subtask transitions, wrong-object manipulation, or ineffective recovery~\cite{beyondmimic,Deepmimic,iwalker,ze2025generalizable}. 
The underlying issue is physical rather than purely semantic: objects move in metric 3D, contact hands or supports, enter containers, and become partially occluded. 
Avoiding object-state divergence therefore requires object-centered state that is persistent over time, actionable by the policy, and verifiable after each executed action chunk.

To provide this state, we propose \emph{Persistent Object Tokenization} (POT). 
POT lifts task-relevant RGB-D evidence into role-indexed 3D object records and serializes these records as Persistent 3D Object Tokens for a whole-body action expert. 
Instantiated as \emph{POT-VLA}, the method closes the perception-action-verification loop over a shared object memory. 
Before each action chunk, the action expert consumes object tokens that expose active task roles, metric 3D locations, relation features, and uncertainty. 
After execution, RGB-D observations refresh the same object records, allowing grounded geometric predicates to verify whether the intended physical relation has been achieved. 
POT does not assume a simulator, explicit physics engine, or learned dynamics model; it records a measurement-grounded stream of object-state transitions, including grasping, displacement, containment, support, and handover progress.

We evaluate whether this shared object-state loop improves real humanoid execution on a Unitree G1 across eight real-world task families spanning transport, object-to-container placement, bimanual handling, stacking, sorting, articulated interaction, deformable-object transport, and close-range handover. 
Against a matched direct baseline with the same embodiment, action expert, and runtime, POT-VLA improves task success from 39/80 to 71/80. 
As an external reference, POT-VLA achieves 44/50 successes on our Being-0-aligned service-task suite, while the Being-0 paper reports 37/50 on its corresponding service tasks.

Our contributions are:
\begin{itemize}
\item We introduce POT, a role-aware object-level physical memory that binds task entities to persistent role slots, refreshes their metric 3D evidence, and serializes identity, geometry, visibility, uncertainty, and spatial relations into Persistent 3D Object Tokens.
\item We insert Persistent 3D Object Tokens into the whole-body action-head sequence, enabling object-state-conditioned action prediction. By exposing target, destination, support, and handover states as explicit 3D tokens, POT-VLA grounds loco-manipulation chunks in task-relevant physical object state rather than relying solely on implicit visual-language features.
\item We build a closed-loop POT-VLA system in which the action expert and predicate verifier share the same refreshed object memory. After each executed action chunk, grounded geometric predicates verify the resulting physical relations and trigger recovery when the desired relation is not achieved.
\end{itemize}

\section{Related Work}

We organize related work around the components required for verifiable humanoid VLA execution: generalist action policies, object-centric 3D grounding, whole-body humanoid execution, and execution verification. 
POT-VLA contributes the coupling representation: refreshed object-level physical memory that is both actionable by the policy and verifiable after execution.

\subsection{Generalist Vision-Language-Action Policies}
Vision-language-action policies map language, visual observations, and robot state to actions at scale~\cite{rt1,rt2,openxembodiment,octo,pi_0,pi05}. 
While these models enable broad visuomotor generalization, task-relevant object state is typically encoded implicitly inside visual-language features, making it difficult to inspect, refresh after contact, or share with a verifier. 
POT-VLA keeps the learned action expert as the primary controller, but exposes target, destination, support, and handover state as Persistent 3D Object Tokens inserted into the action-head sequence.

\subsection{Object-Centric 3D Grounding}
Object-centric grounding methods translate language or visual goals into spatially actionable representations, including prompt-conditioned manipulation, 3D value maps, affordance prediction, and object-centric policies~\cite{vima,voxposer,cliport,peract,rvt,structdiffusion}. 
These methods often provide current targets, affordances, or action-conditioned spatial inputs, whereas humanoid loco-manipulation requires task entities to remain addressable across walking, grasping, occlusion, placement, and recovery. 
POT turns RGB-D grounding into persistent role-indexed object records that are refreshed across action chunks, serialized into object tokens for action prediction, and reused for geometric verification.

\subsection{Humanoid Loco-Manipulation Systems}
Humanoid loco-manipulation couples locomotion, stance selection, reachability, manipulation, and balance~\cite{beyondmimic,Deepmimic,iwalker,ze2025generalizable,radosavovic2024real}. 
Recent humanoid systems combine high-level reasoning with modular skills, intermediate decision layers, or VLM-based monitoring~\cite{being0,Leverb,wang2024autonomous,hierarchicalhumanoidvlp}, while learning-heavy whole-body policies pursue broader generalization~\cite{wholebodyvla,humanoidverse}. 
POT-VLA follows the learned whole-body policy direction, but conditions action chunks and verifies their physical consequences using the same refreshed object records.

\subsection{Execution Verification and Replanning}
Language-conditioned robot agents and verification systems decompose instructions, select feasible skills, generate programs, or use feedback for replanning~\cite{saycan,codeaspolicies,huang2022language,innermonologue,verifyllm,scenegraphreplan,replanvlm,agenticlab,yao2022react}. 
Their feedback state is often symbolic, skill-centric, or separated from the learned action stream, which can create object-state divergence in humanoid service manipulation. 
POT-VLA instead closes the loop over shared object-centric state: after each executed action chunk, RGB-D evidence refreshes the same role-indexed records that conditioned the action, and grounded geometric predicates verify the resulting physical relations.

\section{Methods}

\begin{figure*}[t]
  \centering
  \includegraphics[width=\textwidth]{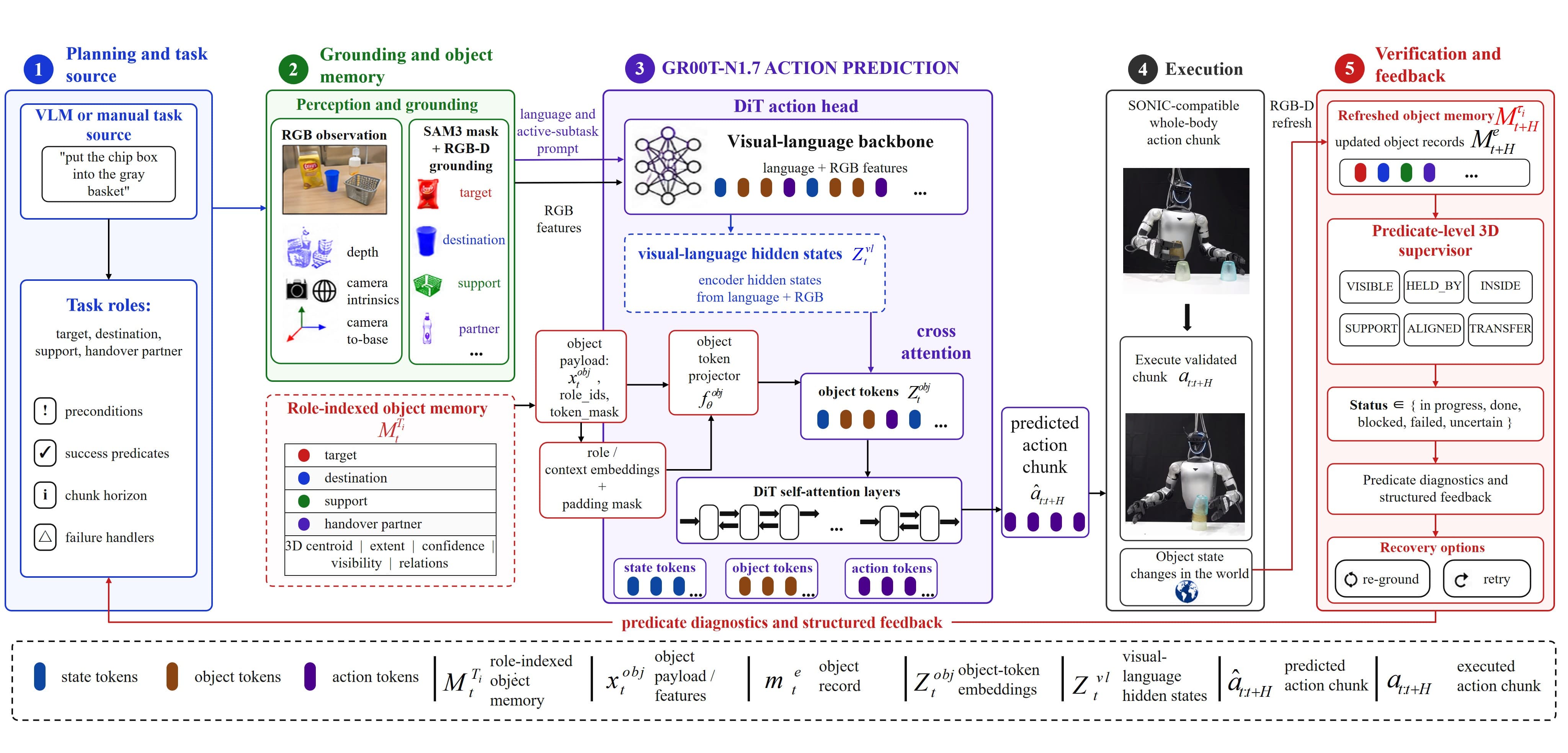}
  \caption{POT-VLA system overview. POT maintains a role-indexed object memory $\mathcal{M}^{\tau_i}_t$, serializes it into Persistent 3D Object Tokens for the whole-body action head, and refreshes the same memory after each executed chunk. The predicate supervisor verifies the refreshed memory $\mathcal{M}^{\tau_i}_{t+H}$ with grounded 3D predicates, closing the maintain--act--observe--verify--recover loop over shared object-level physical state.}
  \label{fig:system}
\end{figure*}

We study closed-loop humanoid loco-manipulation from instruction $I$, RGB-D observations, and proprioception. 
POT-VLA maintains a role-indexed object memory $\mathcal{M}^{\tau_i}_t$ for the active typed subtask $\tau_i$, shared by the action expert and predicate verifier. 
POT builds this memory in the robot base frame and serializes it as Persistent 3D Object Tokens for predicting a whole-body chunk $\hat{a}_{t:t+H}$. 
After runtime validation and execution as $a_{t:t+H}$, RGB-D refresh updates the same memory to $\mathcal{M}^{\tau_i}_{t+H}$ for physical-consistency checks, yielding the ground--act--refresh--verify--recover cycle in Fig.~\ref{fig:system}.

\subsection{Persistent Object Tokenization (POT)}\label{pot}

\paragraph{Task-role object records.}
POT starts from the active typed subtask rather than a particular planner implementation. 
Given instruction $I$ and the current observation context, a VLM planner or human-authored task file provides a typed-subtask plan $\Pi=\langle \texttt{task\_id}, I, \{\tau_i\}_{i=1}^{N}\rangle$. 
Each subtask specifies roles, grounding queries, success predicates, failure handlers, chunk horizon, timeout, and retry budget, but not low-level motions. 
For the active subtask $\tau_i$, let $\mathcal{E}_i$ be its task-relevant entity set. 
Given RGB-D observation $o_t$ and proprioception $s_t$, POT constructs a role-indexed object memory $\mathcal{M}^{\tau_i}_t=\{m_t^e \mid e\in\mathcal{E}_i\}$, where each record $m_t^e$ stores the entity role, phrase, image box, 3D centroid and extent, confidence, visibility, and relation features. 
In our implementation, online or cached SAM3 masks provide role/query masks~\cite{sam3}; valid depth pixels are back-projected through camera intrinsics and the camera-to-base transform, filtered by calibrated workspace bounds, and summarized in the robot base frame. 
The resulting record fields include object--container distance, support height, end-effector--object offset, target--destination displacement, alignment, containment, and handover cues.

\paragraph{Persistent token schema.}
Persistent 3D Object Tokens are the tokenized form of this memory, $x_t^{obj}=T(\mathcal{M}^{\tau_i}_t)$, where $T$ maps object records to a fixed-slot feature tensor. 
The default schema allocates $K=8$ slots and $F=33$ features per slot. 
Valid slots carry role IDs such as \texttt{TARGET}, \texttt{DESTINATION}, \texttt{SUPPORT}, and \texttt{HANDOVER\_PARTNER}. 
Unused slots are \texttt{PADDING} and masked out. 
Slot features include visibility, confidence, normalized 2D box coordinates, 3D centroid, 3D extent, optional orientation fields, and relations. 
Persistence comes from keeping task entities bound to role slots while refreshing their metric evidence after each action chunk.
If an object is occluded or low-confidence, the role slot remains present, but visibility, confidence, and mask fields mark the record as uncertain, allowing the verifier to request re-observation rather than treating the object as confirmed.
At runtime, the action head consumes the payload $(x_t^{obj}, \texttt{role\_ids}, \texttt{token\_mask})$; these tokens do not script end-effector poses or decide task success, but expose the same role-indexed object state to the learned action expert and predicate supervisor.

\subsection{Object-Token Conditioned VLA Execution}\label{exec}

We instantiate the learned action expert with GR00T-N1.7 and keep the same Unitree-G1 whole-body action representation and runtime~\cite{groot}. 
The direct baseline conditions on language, visual observations, and proprioception. 
POT-VLA keeps the same backbone, embodiment, and runtime, but augments the action head with Persistent 3D Object Tokens before each chunk and checks the refreshed object memory after execution. 
Thus the controlled policy-side change is the shared object-state loop rather than a different robot controller or action space.

\paragraph{Role-aware object projection.}
For the active subtask, the processor converts the task-relevant records into $x_t^{obj}\in\mathbb{R}^{K\times F}$. 
Slot $k$ contains $[v_k,q_k,b_k,c_k,s_k,o_k,\rho_k]$, denoting visibility, confidence, 2D box, 3D centroid, 3D extent, optional orientation, and relation features. 
All metric fields are expressed in the robot base frame and normalized before tokenization. 
The object-token projector embeds these slot features into the DiT hidden dimension, $Z_t^{obj}=f_{\theta}^{obj}(x_t^{obj})\in\mathbb{R}^{K\times d}$, where $\theta$ denotes learned action-head parameters. 
In implementation, $f_{\theta}^{obj}$ is a \texttt{LayerNorm} $\rightarrow$ \texttt{Linear} $\rightarrow$ \texttt{GELU} $\rightarrow$ \texttt{Dropout} $\rightarrow$ \texttt{Linear} projection. 
Each projected slot is combined with a learned role embedding and a learned context embedding for object slots, allowing the action head to separate task role semantics from generic object feature content. 
Padded slots are masked, and optional confidence gating scales features using visibility and confidence.

\paragraph{Action-head insertion.}
POT-VLA inserts object tokens into the DiT action-head self-attention sequence rather than into the visual-language backbone. 
The sequence layout is $S_t=[Z_t^{state},Z_t^{obj},Z_t^{action}]$, with padded object slots masked out, while visual-language backbone features remain encoder hidden states for cross-attention. 
Here $Z_t^{state}$, $Z_t^{obj}$, and $Z_t^{action}$ denote state, object, and action tokens, and $Z_t^{vl}$ denotes visual-language hidden states. 
The action expert predicts a candidate whole-body chunk $\hat{a}_{t:t+H}=\pi_{\theta}(S_t,Z_t^{vl})$, where $Z_t^{vl}$ contains language and image features. 
Depending on the active subtask, the chunk can express coordinated navigation, reaching, grasping, carrying, placing, or handover. 
After runtime checks and command smoothing, the executed chunk is denoted $a_{t:t+H}$. 
This shared action representation expresses navigation, reaching, grasping, carrying, placing, and handover within one whole-body action stream, rather than introducing a task-level handoff between locomotion and manipulation policies.

\paragraph{Training and adaptation.}
POT-VLA uses the same action-chunk prediction objective as the base action expert, and no auxiliary object-memory or predicate loss is added. 
During object-token fine-tuning, demonstrations are paired with object-token sidecars produced by the same role-aware RGB-D schema used at deployment, so the action head observes aligned language, proprioception, visual features, and object payloads during training. 
Gradients from the action loss update the object-token branch together with the enabled action-head parameters. 
When object-token conditioning is disabled, the base action-expert training and inference path is unchanged.

\paragraph{Closed-loop execution and safety.}
POT-VLA executes only one validated short-horizon chunk before refreshing perception. 
At $t^+=t+H$, RGB-D observations update the same object memory to $\mathcal{M}^{\tau_i}_{t^+}$, regenerate $Z_{t^+}^{obj}$, and provide the evidence used to continue the current subtask or trigger recovery. 
The humanoid runtime applies command smoothing, joint and velocity limits, balance checks, collision-sensitive workspace constraints, and emergency stops.

\subsection{Predicate-Level Verification and Recovery}\label{sup}

The geometric predicate supervisor is the physical consistency layer of POT-VLA. 
It reads the same role-indexed object memory that conditions the action policy, checks subtask readiness, and verifies the refreshed memory after each action chunk. 
After execution from $t$ to $t^+=t+H$, completion is decided on $\mathcal{M}^{\tau_i}_{t^+}$. 
A subtask is therefore complete only when the intended metric relation holds in the refreshed 3D object state, not simply when the policy attempted the corresponding action.

Let $\Psi_i$ be the set of physical-relation predicates for subtask $\tau_i$, covering containment, support, end-effector--target proximity, target--destination displacement, bimanual assignment, and handover-partner distance. 
Each predicate is represented as $p=\langle \kappa,\alpha,\mathrm{op},\nu,n\rangle$. 
Here $\kappa$ is the predicate type, $\alpha$ denotes grounded object-record arguments, $\mathrm{op}$ is the comparison operator, $\nu$ is the threshold or target value, and $n$ is the required stability window. 
Thresholds and stability windows are specified by the typed subtask and robot calibration, rather than being learned by the action policy. 
At verification index $\xi$, scalar evidence $\phi(\kappa,\alpha,\mathcal{M}^{\tau_i}_{\xi})$ determines whether the predicate is satisfied. 
Invisible, low-confidence, or unstable records yield \texttt{uncertain} unless safety is violated.

The supervisor evaluates readiness using preconditions $\mathcal{P}_i\subseteq\Psi_i$ and completion using goals $\mathcal{G}_i\subseteq\Psi_i$, returning one of \{\texttt{in\_progress}, \texttt{done}, \texttt{blocked}, \texttt{failed}, \texttt{uncertain}\}.
Precondition failures produce \texttt{blocked}, satisfied goals produce \texttt{done}, and timeout, safety, or feasibility violations produce \texttt{failed}.
Low-confidence or unstable object evidence produces \texttt{uncertain} and triggers re-observation or re-grounding; otherwise, the subtask remains \texttt{in\_progress} and the action expert receives another chunk request with refreshed object tokens.

The supervisor also returns diagnostics $\mathbf{d}_{\xi}$ that localize the failure mode, such as missing visibility, failed grasping, unstable support, incorrect placement, or excessive handover distance. 
For semantic conditions not fully captured by geometry, such as object-category ambiguity or handover acceptance, the system can optionally query a VLM verifier, while reliable geometric predicates take precedence in our reported experiments. 
Based on the status and diagnostics, POT-VLA chooses among \texttt{continue}, \texttt{retry}, \texttt{reobserve}, \texttt{reground}, and \texttt{replan}. 
Local recovery refreshes or regrounds object records before task-level replanning receives compact feedback such as \texttt{object\_not\_grasped} or \texttt{outside\_goal\_region}.


\section{Experiments}

Our experiments evaluate whether the shared object-state loop improves real-world humanoid execution, identify the contribution of object-token conditioning and predicate-level verification, and test robustness under object-state changes including novel instances, shifted layouts, distractors, and mid-execution perturbations.

\subsection{Setup and Protocol}\label{sec:setup}
\begin{figure}[t]
  \centering

  \noindent
  \taskblock
    {Task 1: Cart transport and placement}
    {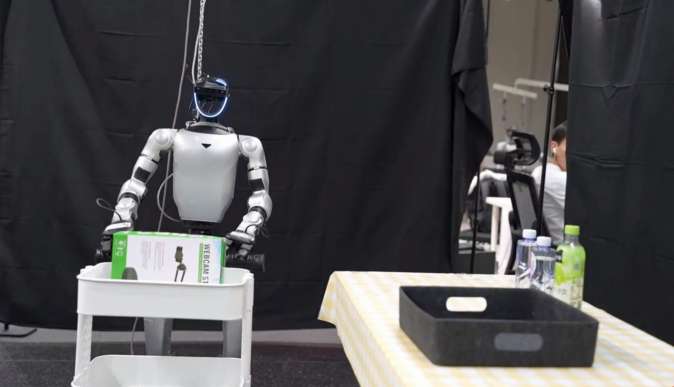}
    {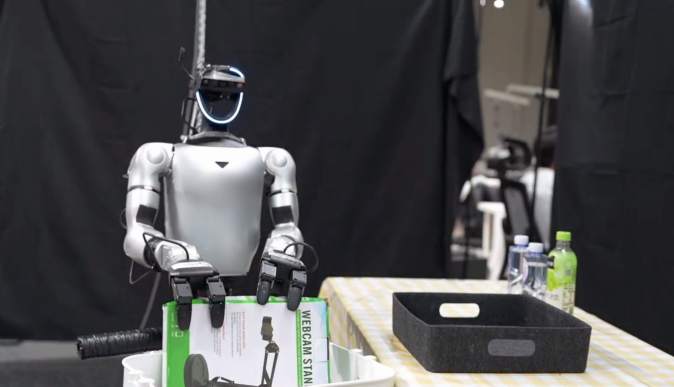}
    {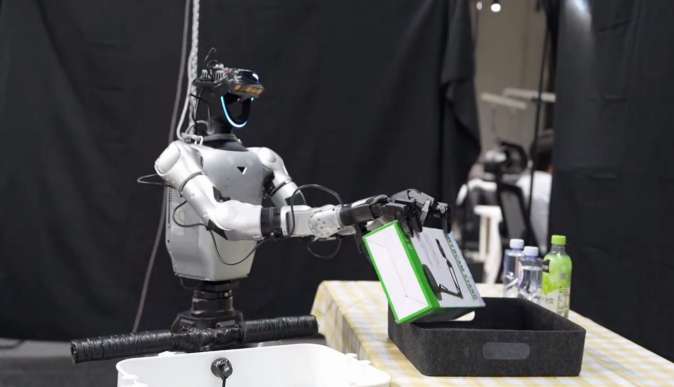}
  \hfill%
  \taskblock
    {Task 2: Chip box into gray basket}
    {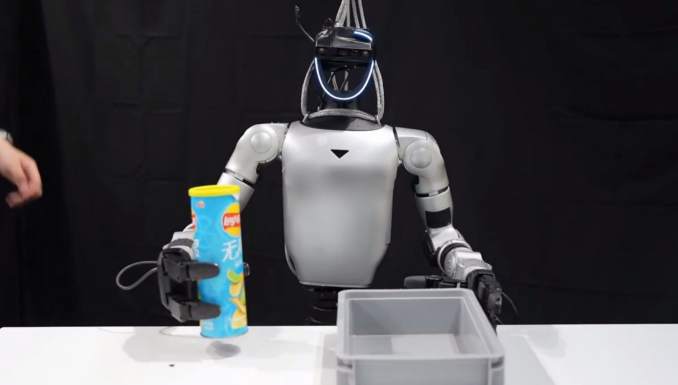}
    {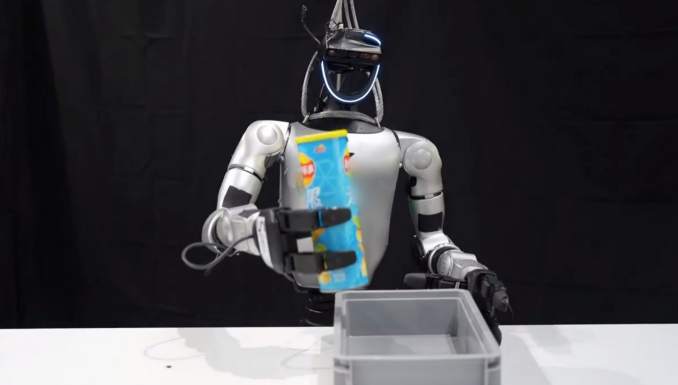}
    {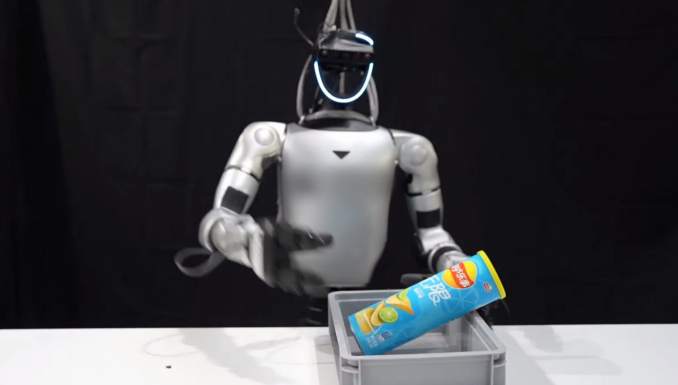}

  \par\vspace{3pt}

  \noindent
  \taskblock
    {Task 3: Two balls into gray basket}
    {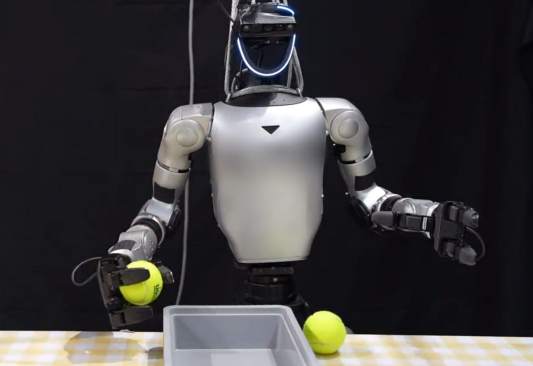}
    {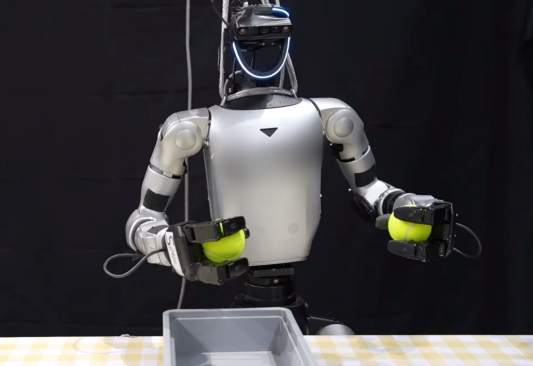}
    {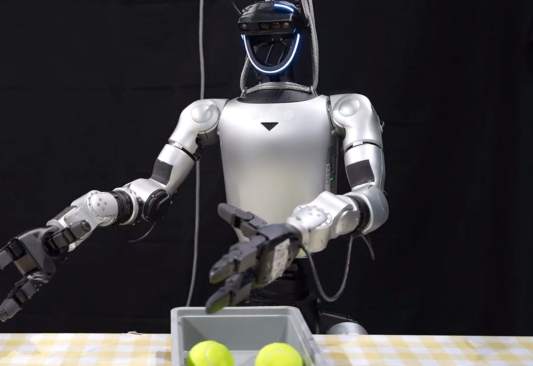}
  \hfill%
  \taskblock
    {Task 4: Stack three cups}
    {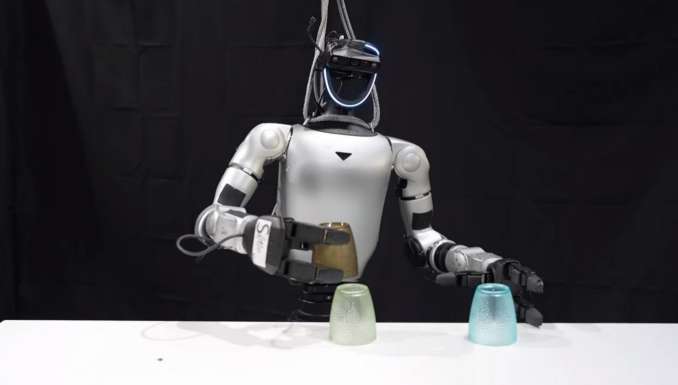}
    {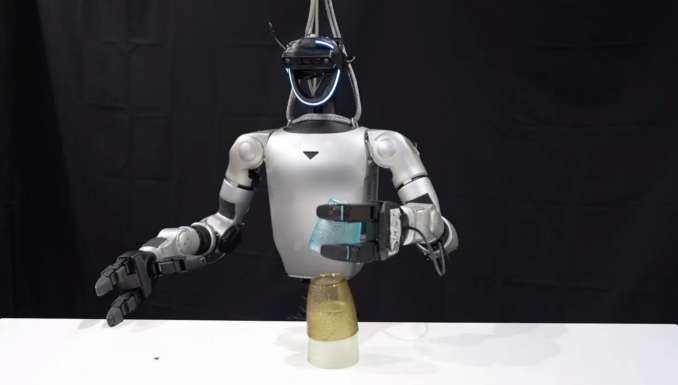}
    {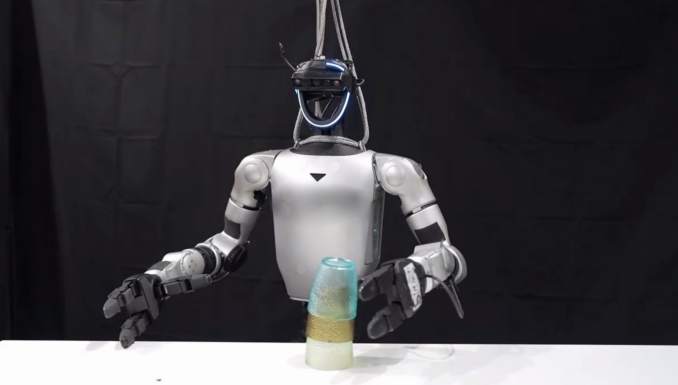}

  \par\vspace{3pt}

  \noindent
  \taskblock
    {Task 5: Garments to laundry basket}
    {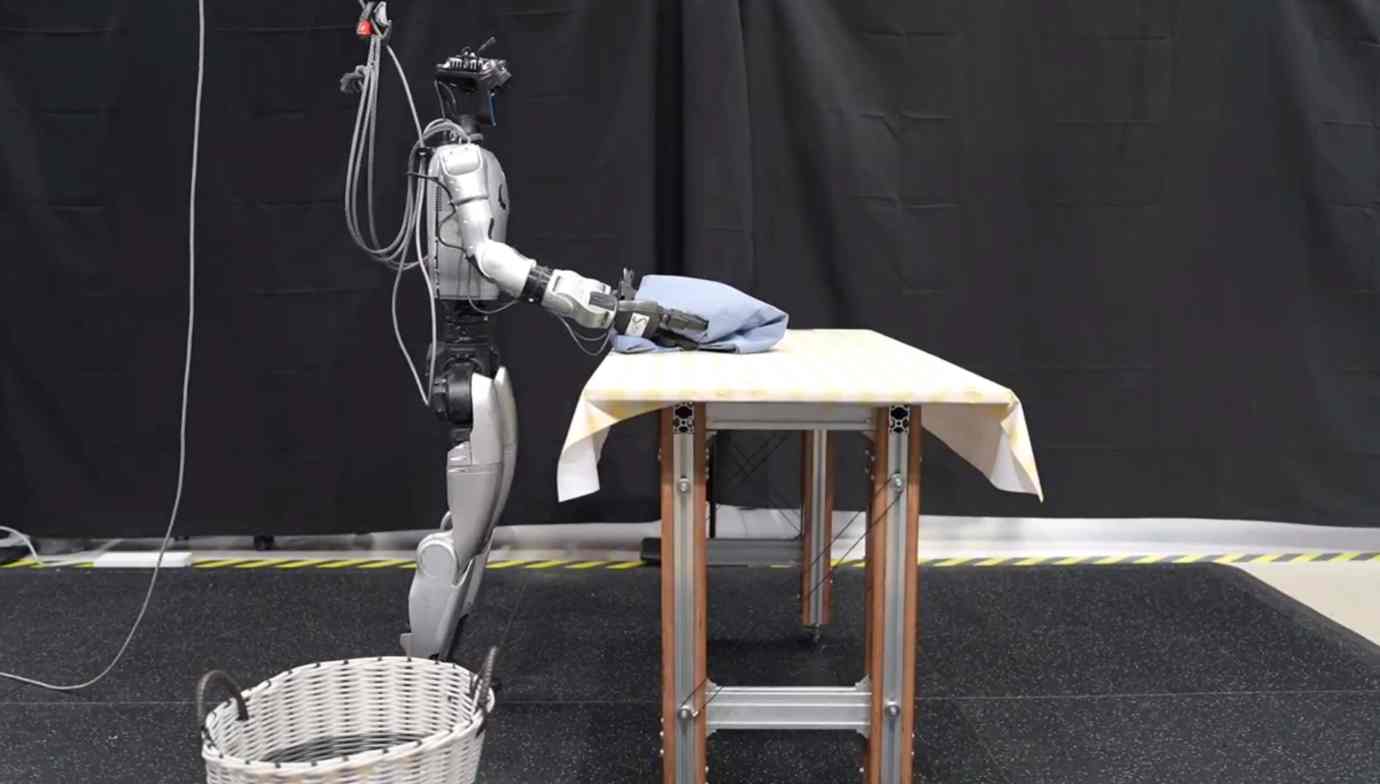}
    {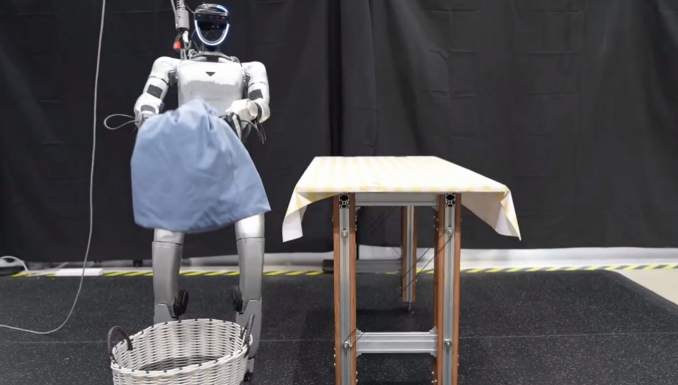}
    {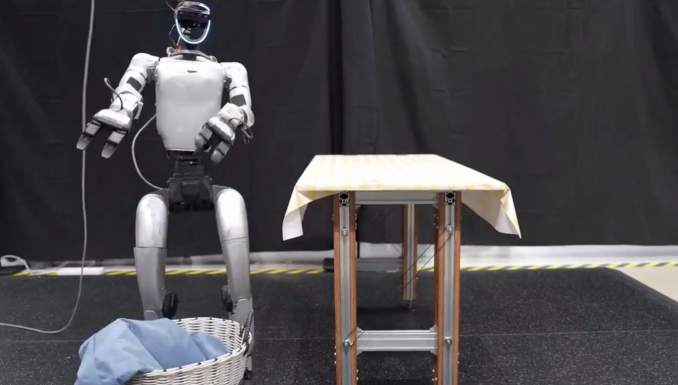}
  \hfill%
  \taskblock
    {Task 6: Drawer/tray place-and-close}
    {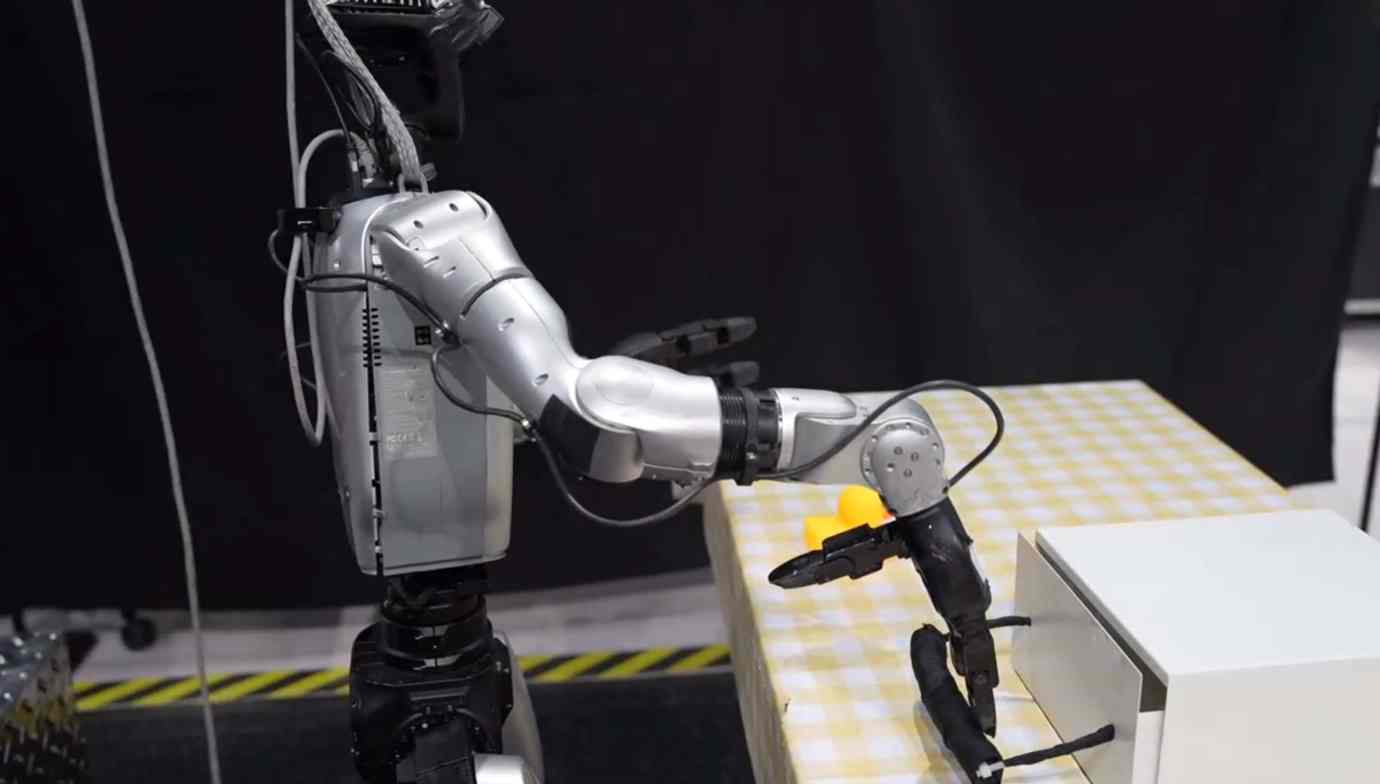}
    {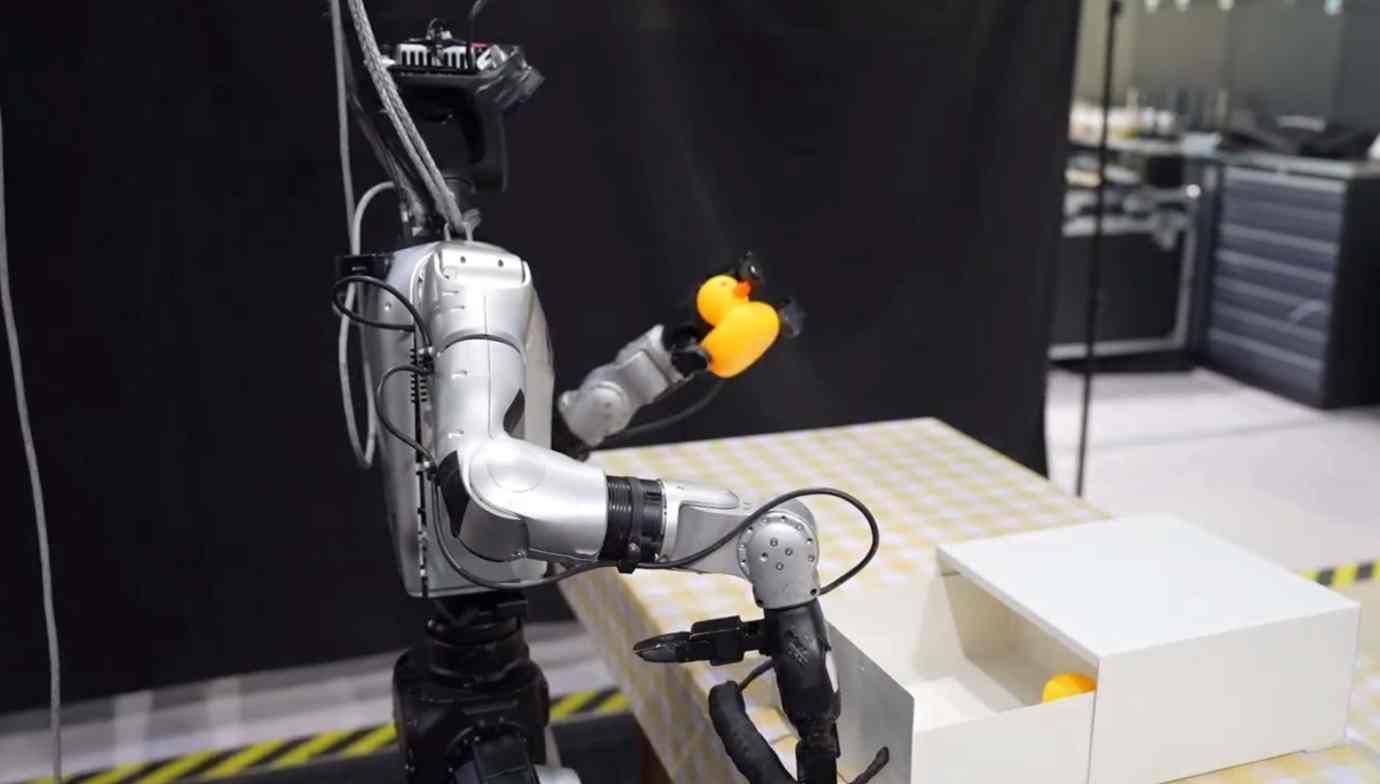}
    {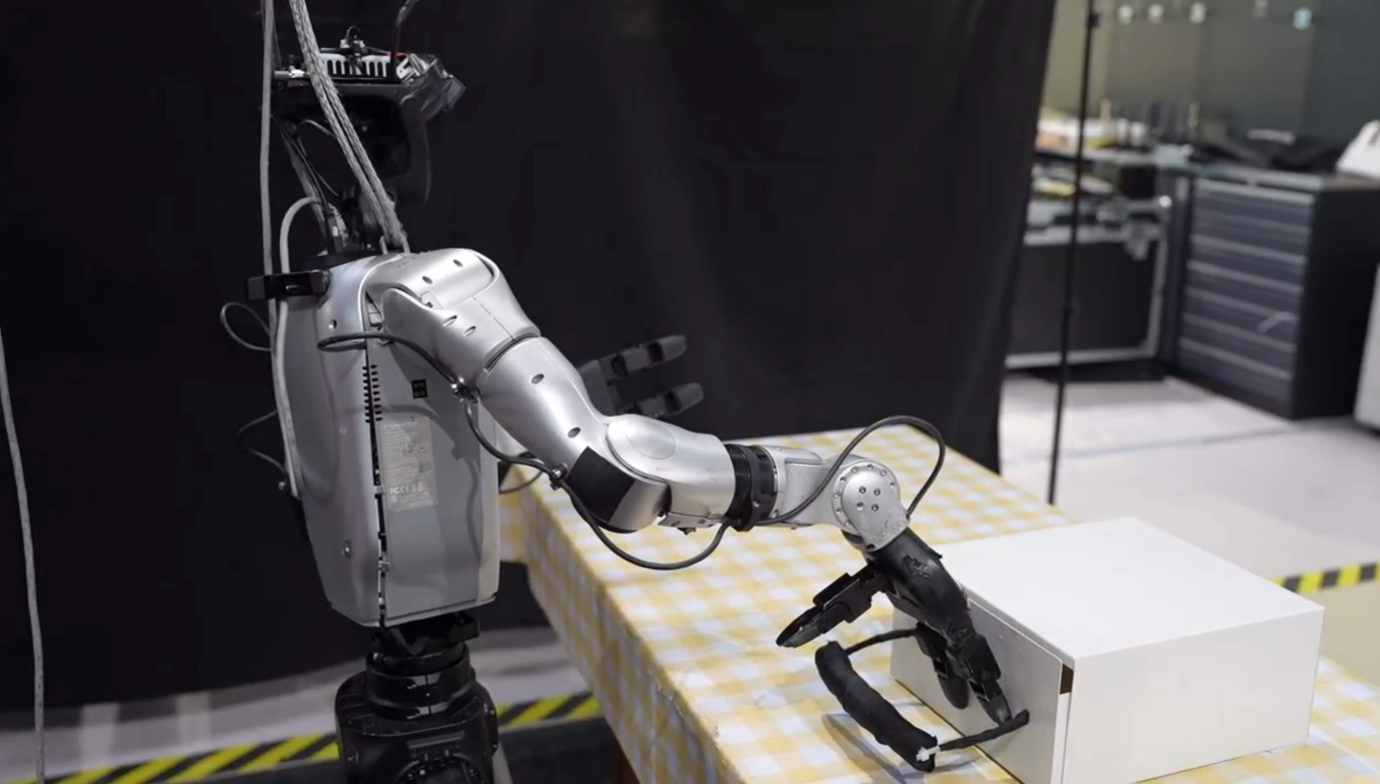}

  \par\vspace{3pt}

  \noindent
  \taskblock
    {Task 7: Tabletop sorting}
    {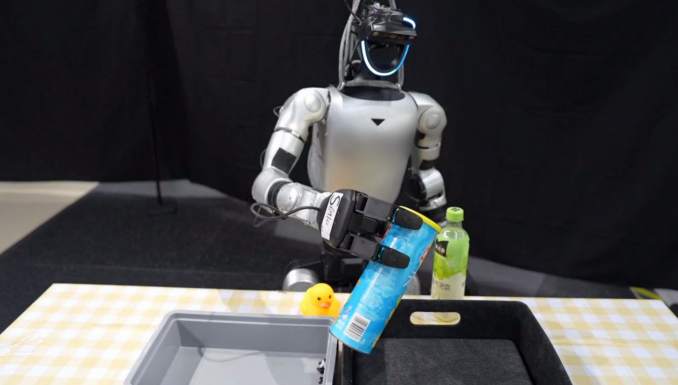}
    {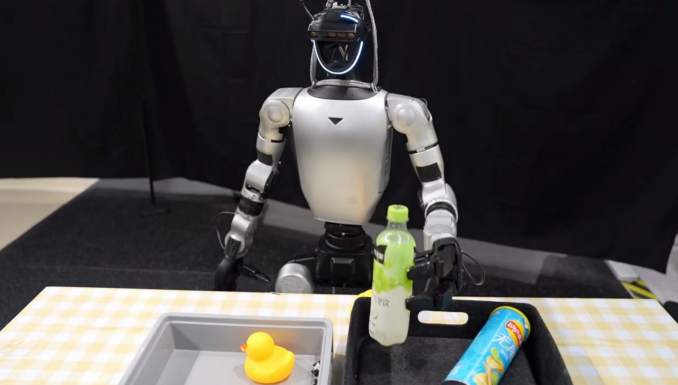}
    {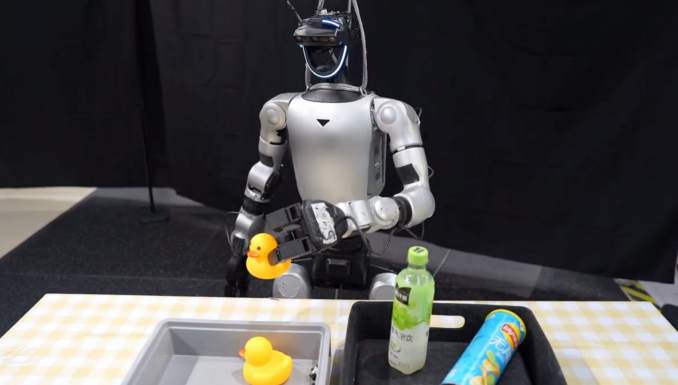}
  \hfill%
  \taskblock
    {Task 8: Close-range drink handover}
    {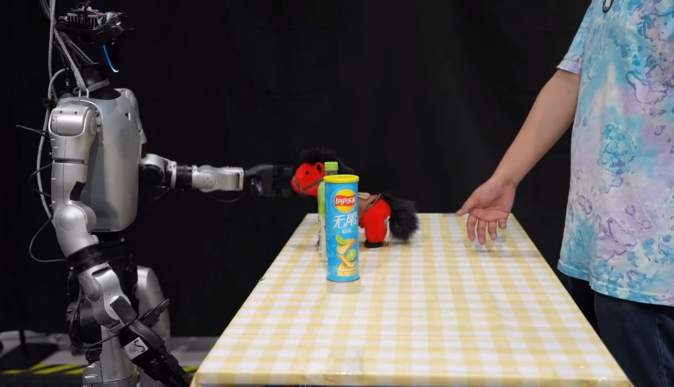}
    {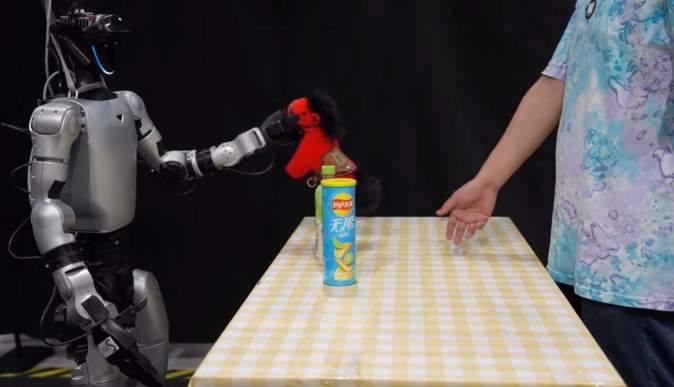}
    {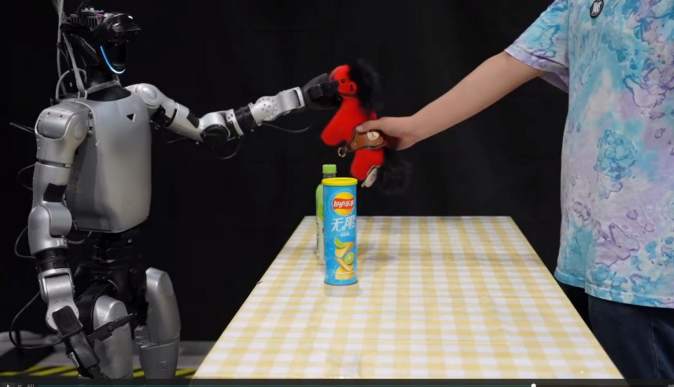}

  \caption{
    Real-world task rollouts. Each task family is shown with three
    representative frames from the same execution sequence, illustrating
    the object-centered state changes that POT-VLA conditions on and
    verifies across transport, placement, stacking, sorting, articulated
    interaction, deformable-object handling, and handover tasks.
  }
  \label{fig:task_rollouts}
\end{figure}

We evaluate POT-VLA on a Unitree G1 humanoid with a Dex3-1 dexterous hand and a head-mounted RGB-D camera in a cluttered office environment. 
A desktop workstation runs perception, typed-subtask execution, object-token projection, and predicate supervision, while the onboard Unitree computer runs low-level arm and locomotion controllers. 
Each reported entry uses 10 real-world trials unless stated otherwise. 
A trial succeeds only if the final geometric predicates remain satisfied over the required temporal window within the predefined timeout and retry budget, without human intervention. 
For Being-0, we use the success counts reported in the Being-0 paper on overlapping service tasks. 
This reference is not a local reproduction. 
For the policy backbone, POT-VLA and the direct baseline use the same GR00T-N1.7 action expert, embodiment, runtime, and equivalent task-language prompt. 
The direct baseline disables persistent object records, object tokens, and geometric predicate supervision during execution. 
It follows the same instruction-level execution schedule, but does not use predicate feedback for local recovery.
We report the direct baseline both in the main comparison and as the no-token, no-verifier variant in the ablation.
Table~\ref{tab:task_suite} summarizes the real-world task families. 
Fig.~\ref{fig:task_rollouts} shows representative three-frame rollouts for all eight task families.
Appendix~\ref{app:experiments} provides additional token-field details.

\subsection{Main Real-World Results}\label{sec:lh_results}
Table~\ref{tab:real_world_results} summarizes the main results, separating the matched direct-baseline comparison from the external literature reference. 
Across the eight-task suite, POT-VLA completes 71/80 trials compared with 39/80 for the direct baseline. 
As an external reference, POT-VLA completes 44/50 trials on our Being-0-aligned service-task suite, while the Being-0 paper reports 37/50 on its corresponding service-task suite.
The matched comparison supports the state-divergence hypothesis. 
The largest gaps appear in tasks that require maintaining object relations across multiple chunks, such as cup stacking, two-ball placement, garment transport, and drawer/tray interaction. 
The direct baseline must infer the active target, destination, support, and handover partner implicitly from language-image features and proprioception. 
POT-VLA instead provides the same action expert with role-aware 3D object tokens before each action chunk, then checks the refreshed object memory with geometric predicates before advancing. 
In our rollouts, many direct-baseline failures are not semantic misunderstandings of the instruction, but metric object-state errors: inaccurate grasp targets on cart handles or carried objects, misaligned cups during stacking, missed small objects during sorting, failed ball grasps, or placements that land outside the intended container. 
In POT-VLA, missed grasps, outside-goal placements, and false handover completions trigger re-observation, re-grounding, retry, or replanning rather than silent task progression.

\begin{table*}[t]
\centering
\small
\setlength{\tabcolsep}{4pt}
\renewcommand{\arraystretch}{1.03}
\caption{Real-world evaluation results. Entries are successes/trials. Panel A is the primary matched comparison against the direct baseline under the same action backbone, embodiment, and execution conditions. Panel B is an external paper-reported reference on Being-0-aligned service tasks, not a local reproduction of Being-0.}
\label{tab:real_world_results}
\begin{minipage}[t]{0.62\textwidth}
\centering
\footnotesize
\textbf{A. Matched direct baseline comparison}\\[-2pt]
\begin{tabular}{@{}lcc@{}}
\toprule
\textbf{Task} & \textbf{Direct} & \textbf{POT-VLA} \\
\midrule
Cart transport/place & 3/10 & \textbf{8/10} \\
Chip box to basket & 9/10 & \textbf{10/10} \\
Two balls to basket & 5/10 & \textbf{9/10} \\
Stack three cups & 1/10 & \textbf{8/10} \\
Garments to basket & 3/10 & \textbf{9/10} \\
Drawer/tray place-close & 4/10 & \textbf{8/10} \\
Tabletop sorting & 5/10 & \textbf{9/10} \\
Close-range handover & 9/10 & \textbf{10/10} \\
\midrule
\textbf{Total} & 39/80 & \textbf{71/80} \\
\bottomrule
\end{tabular}
\end{minipage}\hfill
\begin{minipage}[t]{0.34\textwidth}
\centering
\footnotesize
\textbf{B. External Being-0-aligned reference}\\[-2pt]
\begin{tabular}{@{}lcc@{}}
\toprule
\textbf{Task} & \textbf{Being-0} & \textbf{POT-VLA} \\
\midrule
Fetch-bottle & 9/10 & 9/10 \\
Deliver-basket & 8/10 & 8/10 \\
Grasp-bottle & 8/10 & \textbf{10/10} \\
Place-basket & 6/10 & \textbf{9/10} \\
Place-coffee & 6/10 & \textbf{8/10} \\
\midrule
\textbf{Total} & 37/50 & \textbf{44/50} \\
\bottomrule
\end{tabular}
\end{minipage}
\end{table*}

\subsection{Ablations and Object-State Generalization}\label{sec:ablation_generalization}
We use ablations and generalization tests to separate the effect of object-token conditioning from closed-loop verification. 
All variants keep the Unitree G1 embodiment, action representation, runtime checks, and task-language context fixed. 
The ablation variants differ only in whether object-token conditioning and predicate-level recovery are enabled during execution. 
The direct and verifier-only variants use the same direct action-expert checkpoint, while the POT-token variants use the same object-token fine-tuned checkpoint, demonstration sidecars, fine-tuning budget, and checkpoint-selection rule. 
The ablations remove one part of the shared object-state loop at a time: \emph{Verifier only} keeps the direct action path but adds predicate-level checks and local recovery, while \emph{POT tokens only} conditions the action head on Persistent 3D Object Tokens but disables predicate-level recovery. 
The generalization tests perturb the object state without changing the task family.

\begin{table*}[t]
\centering
\small
\setlength{\tabcolsep}{4pt}
\renewcommand{\arraystretch}{1.05}
\caption{Ablation and generalization results under the same Unitree G1 execution setup. Entries are successes/trials. Panel A isolates object-token conditioning and predicate-level verification on four representative tasks, with 40 trials per variant. Panel B evaluates robustness under object-state shifts, with 10 matched trials per setting.}
\label{tab:ablation_generalization}
\begin{minipage}[t]{0.48\textwidth}
\centering
\footnotesize
\textbf{A. Shared object-state loop ablations}\\[-2pt]
\begin{tabular}{@{}lccc@{}}
\toprule
\textbf{Variant} & \textbf{Tokens} & \textbf{Verifier} & \textbf{Success} \\
\midrule
Direct baseline & -- & -- & 15/40 \\
Verifier only & -- & \checkmark & 22/40 \\
POT tokens only & \checkmark & -- & 31/40 \\
POT-VLA & \checkmark & \checkmark & \textbf{34/40} \\
\bottomrule
\end{tabular}
\end{minipage}\hfill
\begin{minipage}[t]{0.48\textwidth}
\centering
\footnotesize
\textbf{B. Generalization under object-state shifts}\\[-2pt]
\begin{tabular}{@{}lcc@{}}
\toprule
\textbf{Shift} & \textbf{Direct} & \textbf{POT-VLA} \\
\midrule
Novel object instances & 6/10 & 9/10 \\
Shifted object poses/layouts & 5/10 & 9/10 \\
Distractor objects & 8/10 & 9/10 \\
Mid-execution perturbations & 4/10 & 8/10 \\
\bottomrule
\end{tabular}
\end{minipage}
\end{table*}

The ablation subset contains four tasks with large state-divergence gaps in Table~\ref{tab:real_world_results}: two-ball placement, cup stacking, drawer/tray interaction, and tabletop sorting. 
POT tokens provide the largest gain, improving direct execution from 15/40 to 31/40 by giving the action expert explicit role-indexed 3D state before each chunk. 
Verification alone improves success to 22/40 by rejecting false completions and triggering local re-observation or retry. 
Full POT-VLA reaches 34/40, indicating that predicate-level verification provides residual gains on top of the larger improvement from object-token conditioning. 
This pattern matches the intended division of labor: object tokens resolve many grounding failures before action, while verification mainly catches residual false completions and partial failures.
Each generalization setting uses 10 matched trials with controlled object-state shifts. 
For novel instances, we replace task-role objects with unseen items that vary category, appearance, geometry, or grasp affordance, including bottles, cans, cups, balls, and garments. 
For layout shifts, target objects are displaced by 20--50 cm on the tabletop, goal containers or goal regions by 20--60 cm, and object orientations by 30--90 degrees. 
Distractor trials add visually or functionally similar objects near the workspace on top of the natural office clutter. 
Incidental background motion is retained when it occurs. 
Perturbation trials move the target slightly before grasping or after a partial action chunk, requiring the system to refresh object records before continuing. 
The largest generalization gaps appear under novel object instances, shifted object layouts, and mid-execution perturbations, while distractor objects affect both systems less strongly.

Qualitatively, role-aware 3D object tokens reduce target-search oscillation and improve contact-relevant localization for handles, cups, balls, small tabletop objects, garments, and container regions. 
Persistent object records reduce per-chunk identity jitter, and the geometric supervisor refreshes the current subtask locally instead of forcing a full language-level replan. 
These qualitative trends are visible in the rollout frames in Fig.~\ref{fig:task_rollouts}.

\section{Conclusion and Limitations}

We presented POT and \emph{POT-VLA}, a closed-loop humanoid VLA system built around Persistent 3D Object Tokens.
POT turns 3D grounding into execution memory: shared role-indexed object records condition whole-body action generation and verify the physical consequences of executed chunks.
On real Unitree G1 loco-manipulation tasks, this shared object state improves target-directed execution, reduces false success, and enables structured recovery.
POT-VLA remains limited by object-record quality, SAM3/RGB-D reliability, camera calibration, and action-expert embodiment coverage.
Future work will extend POT to active multi-view memory, richer contact, dexterity, and human interaction.

\clearpage
\appendix
\renewcommand{\thesection}{Appendix~\Alph{section}}
\section{Additional Method and Experiment Details}
\label{app:experiments}

In our processor implementation, the object payload enters through \texttt{observation["extras"]["object\_tokens"]}. 
It contains \texttt{x\_t\_obj}, \texttt{token\_mask}, \texttt{role\_ids}, and metadata, and is materialized as \texttt{object\_x\_t\_obj}, \texttt{object\_token\_mask}, and \texttt{object\_role\_ids} before the action head consumes it.

\begin{table*}[h]
\centering
\small
\setlength{\tabcolsep}{5pt}
\renewcommand{\arraystretch}{1.12}
\caption{Object-token representation used by POT-VLA. The same 3D object-record fields condition the VLA action expert and provide predicate evidence for supervision, which keeps action generation and verification grounded in the same object state.}
\label{tab:object_token_representation}
\resizebox{\textwidth}{!}{%
\begin{tabular}{@{}p{0.18\textwidth}p{0.28\textwidth}p{0.24\textwidth}p{0.24\textwidth}@{}}
\toprule
\textbf{Token field group} & \textbf{Examples} & \textbf{VLA execution use} & \textbf{Geometric supervision use} \\
\midrule
Semantic role & \texttt{TARGET}, \texttt{DESTINATION}, \texttt{REFERENCE}, \texttt{SUPPORT}, \texttt{HANDOVER\_PARTNER}, \texttt{PADDING} & Conditions which object to reach, carry, place, sort, stack, or hand over. & Selects predicate arguments for grasped, contained, supported, sorted, or transferred conditions. \\
Grounding evidence & Entity phrase, 2D box, mask-derived proposal confidence, visibility, token mask. & Distinguishes object instances and masks unused slots. & Marks predicates uncertain when required records are invisible or low-confidence. \\
3D geometry & Centroid, extent, optional orientation, support height, container region. & Provides target-directed 3D context without scripting end-effector poses. & Measures distances, containment, support, alignment, and handover transfer. \\
Relational features & End-effector--object offset, target--destination displacement, object--support relation, partner distance. & Conditions reaching, grasping, placing, bimanual assignment, and handover approach. & Localizes failures such as failed grasp, outside-goal placement, unstable support, or excessive handover distance. \\
\bottomrule
\end{tabular}%
}
\end{table*}


\clearpage
\acknowledgments{}


\bibliography{references}  

\end{document}